\documentclass{article}

\usepackage[final,nonatbib]{nips_2018}

\usepackage[utf8]{inputenc} 
\usepackage[T1]{fontenc}    
\usepackage{hyperref}       
\usepackage{url}            
\usepackage{booktabs}       
\usepackage{amsfonts}       
\usepackage{nicefrac}       
\usepackage{microtype}      

\usepackage[pdftex]{graphicx} 
\usepackage{todonotes}
\usepackage[numbers]{natbib}

\title{Advance Prediction of Ventricular Tachyarrhythmias using Patient Metadata and Multi-Task Networks}

\author{Marek Rei$^{\clubsuit\diamondsuit}$ ~ Joshua Oppenheimer$^{\clubsuit\spadesuit}$ ~ Marek Sirendi$^{\clubsuit}$ \\
$^\clubsuit$Transformative AI, New York, United States \\
$^\diamondsuit$Computer Laboratory, University of Cambridge, United Kingdom \\
$^\spadesuit$George Washington University Hospital, Washington D.C., United States \\
{ \small \tt
marek.rei@cl.cam.ac.uk, oppenheimer@transformative.ai, sirendi@transformative.ai
}
}

\begin{document}

\maketitle

\begin{abstract}
We describe a novel neural network architecture for the prediction of ventricular tachyarrhythmias. The model receives input features that capture
the change in RR intervals and ectopic beats, along with
features based on heart rate variability and frequency analysis. 
Patient age is also 
included 
as a trainable embedding, while the whole network is optimized with multi-task objectives.
Each of these modifications provides a consistent improvement to the model performance, achieving 74.02\% prediction accuracy and 77.22\% specificity 60 seconds in advance of the episode.
\end{abstract}

\section{Introduction}


Numerous patients suffer in-hospital cardiac arrest each year, estimated at 209,000 people in the United States alone, of whom only one-in-four survive \cite{merchant2011incidence,heartstatistics}. Cardiac arrest occurs when the heart ceases to
pump blood to the brain and other vital organs. 
20.7\% of cardiac arrests in hospitalized patients occur due to ventricular fibrillation (VF) or pulseless ventricular tachycardia (VT), collectively referred to as ventricular tachyarrhythmia (VTA), as opposed to asystole or pulseless
electrical activity \cite{girotra2012trends}.
Both VT and VF are deadly cardiac
rhythms that must be corrected for the patient to survive. In ventricular tachycardia, the ventricles
beat so quickly that the heart is not able to effectively fill its chambers and pump blood to the
body. 
In ventricular fibrillation,
the ventricles quiver, rather than pump, requiring abortive treatment. 
If either condition is
sustained, the brain dies from the lack of blood flow and oxygen deprivation. In both cases, high-quality CPR can temporarily sustain blood flow to the brain while an electric shock and medications are administered in an attempt to restart
a viable cardiac rhythm.

Patients in critical care units, operating rooms, and telemetry-enabled medical wards undergo
continuous ECG monitoring. 
We hypothesize that a predictive monitoring system utilizing an accurate VTA
prediction algorithm could analyze this signal and alert
healthcare providers when a patient is at imminent risk of developing a ventricular
tachyarrhythmia, improving survival by decreasing the time between onset of VTA and initiation of CPR  and other treatments.
This hypothesis is consistent with research showing that
early defibrillation within the first 2 minutes following in-hospital sudden cardiac arrest is strongly
associated with improved survival-to-hospital-discharge, compared to patients in whom
defibrillation is delayed \cite{chan2008delayed}.  Prevention of sudden cardiac arrest could also be achieved for
patients in the pre-ventricular tachyarrhythmia state, as identified by the algorithm, 
when healthcare providers treat reversible triggers of ventricular arrhythmias \cite{gettes1992electrolyte,wyse2001life}.

In this work, we construct a neural architecture for VTA prediction and investigate novel features based on heart rate variability, combined with patient-level metadata and multi-task network optimization.
Previous research has largely focused on using ECG signals to \textit{detect} various cardiac arrhythmias as soon as they occur \cite{artis1991detection,fokkenrood2007ventricular}.
 The latest work in the area uses deep convolutional neural networks for classifying arrhythmias, achieving performance comparable to human cardiologists \cite{Rajpurkar2017}.
 The \textit{prediction}, as opposed to detection, of ventricular tachyarrhythmias has also been investigated.
Joo et al. \cite{joo2012prediction} describe a neural network for predicting VTA 10 seconds in advance, based on heart rate variability features, evaluating on the  PhysioNet {\small MVTDB} \cite{goldberger2000physiobank} dataset.
Murukesan et al. \cite{Murukesan2014} explored the task of predicting sudden cardiac arrest based on heart rate variability and identified 7 useful features.
Martinez-Alanis et al. \cite{Martinez-Alanis2016} investigated the predictive power of premature ventricular complexes with respect to ventricular tachyarrhythmias.
Lee et al. \cite{lee2016prediction} describe a system for predicting potentially fatal ventricular tachycardias up to an hour in advance. However, their best configuration relies on respiratory features, which can be difficult to obtain, and evaluation is performed only on a private dataset.

\begin{figure*}[t]
	\centering
	\includegraphics[width=1.0\linewidth]{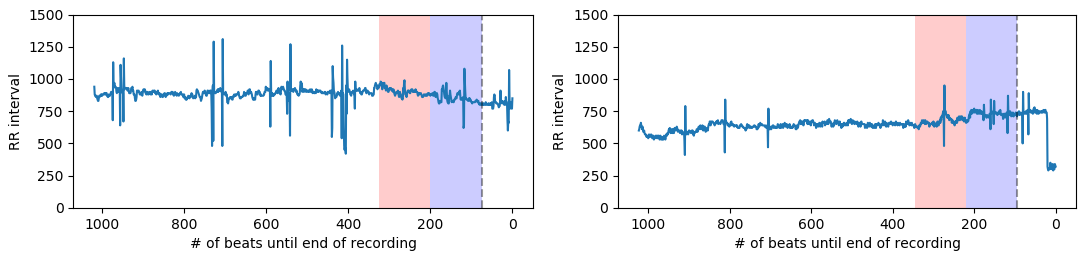}
	\caption{Examples of RR interval sequences in the dataset. A control sample (left) and an episode ending in VTA (right). The dashed line indicates the 60 second boundary; for the prediction task the model does not have access to the signal to the right of this boundary. The red and blue areas indicate the consecutive areas used for calculating the windowed feature values.}
    \label{fig:examples}
\end{figure*}

\section{VTA Prediction Model}
\label{sec:model}

The VTA prediction system receives a sequence of RR intervals\footnote{The time between two consecutive peaks of the QRS complex in an ECG is called the RR interval.} as input and learns to predict whether the monitored patient will experience a ventricular tachyarrhythmia 60 seconds after the end of the available sequence. RR intervals are commonly represented through a collection of metrics based on heart rate variability analysis. Previous work \cite{joo2012prediction,lee2016prediction} has identified 11 features for this purpose, including metrics from the time domain, the frequency domain and from nonlinear analysis. 
From this set, we choose to include three features: 1) the mean of RR intervals, 2) signal power in the low frequency range (0.04–0.15 Hz) and 3) signal power in the high frequency range (0.15–0.4 Hz).
While previous work calculated these features across a 5-minute window, we instead use the values calculated only over the last 30 beats before the decision boundary. This gives the model information about the most recent heart rate without mixing in information from a larger time period.
We also include the minimum and maximum RR interval values over the most recent 30 beats.
Following previous work, ectopic beats were removed before caculating these features.

\begin{figure*}[t]
	\centering
	\includegraphics[width=0.32\linewidth]{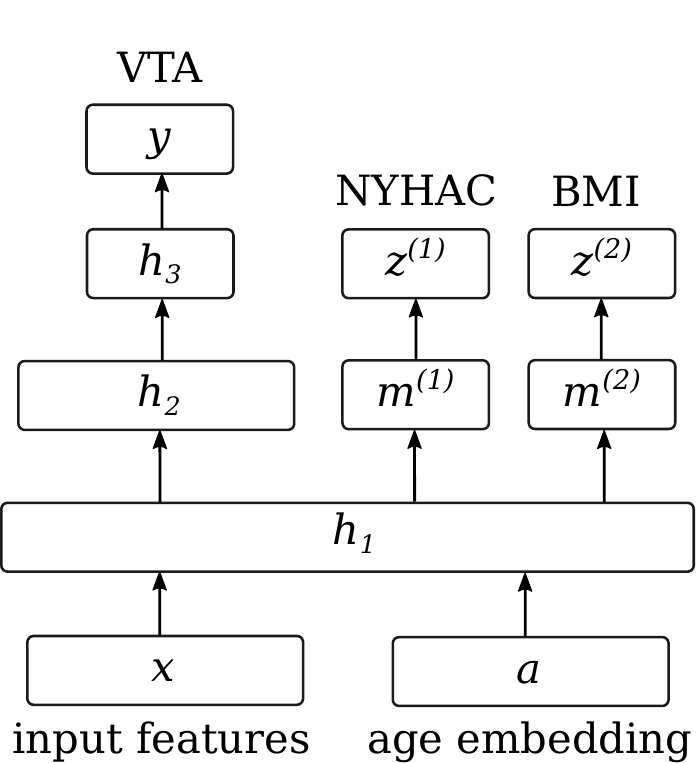}
	\caption{Neural network architecture for VTA prediction. The architecture takes a set of features and an embedding for the patient as input, shares a hidden layer, and predicts the VTA along with NYHAC and BMI.}
    \label{fig:graph}
\end{figure*}

In addition, the model uses novel features based on shifting windows and patient metadata, which we describe in Sections \ref{sec:windowed} and \ref{sec:input}. The final features are range-standardized and given as input to a multi-layer neural network. The architecture has 3 hidden layers with \textit{tanh} activation, containing 150, 100 and 10 neurons. The top of the network has a softmax layer and the model is trained to predict VTA by minimizing cross entropy. Optimization is performed using AdaDelta \cite{Zeiler2012} with the default learning rate $1.0$. Gradients are clipped to $0.1$ and training is performed for 1,000 epochs -- we found that clipping to a very low value but training for a longer period helped the model converge to a more stable solution. Dropout \cite{Srivastava2014a} was applied to the input layer and all hidden layers during training with 75\% probability of keeping the value. The final network, including additional features and multi-task outputs, can be seen in Figure \ref{fig:graph}.

\subsection{Windowed Features}
\label{sec:windowed}

While the previous features describe the sequence of RR intervals, they do not capture the change or directionality in those statistics.
Therefore, we calculate additional feature values across two consecutive windows and use their difference as a new feature for the model, which allows the network to obtain information about the speed and direction in which the values are changing. 
The model includes two windowed features, representing the change in mean RR intervals and the number of ectopic beats. These are calculated over the last 250 heartbeats before the decision boundary, with the split in the middle. For example, the ectopic beats are counted over the most recent 0 to 124 beats, and again over 125 to 250 beats, with the difference between the two values added as a new feature. Figure \ref{fig:examples} illustrates the decision boundary
along with the two consecutive windows that are used for calculating these features. As can be seen, the examples of non-tachyarrhythmic recordings (Fig. \ref{fig:examples}, left) also contain ectopic beats, which makes the task and dataset very challenging. However, the relative increase in ectopic beats is higher for the tachyarrhythmic example (Fig. \ref{fig:examples}, right).

\subsection{Patient Age Embedding}
\label{sec:input}

Previous work has only focused on capturing information directly from the recorded RR intervals or other sensors. However, prior knowledge about the patient can also help the algorithm make more accurate and personalized decisions.
In this model, we experiment with one patient-level input feature -- their age. Particularly, we encode their year of birth and round it to the nearest decade.
Each decade is assigned to a trainable embedding of length 10, which is initialized randomly and updated during optimization -- we found that this representation gave the model more flexibility and also helped manage instances where the patient birth year was unknown. 
The correct embedding for each patient is concatenated to the list of other features and used as input to the network. The embeddings are continuously updated during training;
during testing, the fixed pre-trained embedding values are used.
While it is likely that patient age also correlates with the general probability of experiencing arrhythmias, in this case the model is using the feature for more intricate information -- since the positive and negative examples for each age group are balanced, the benefit in performance comes from the model using the age information to better condition the analysis of the other features.

\subsection{Multi-Task Optimization}
\label{sec:aux}

The regular network is only optimized with cross entropy for predicting the correct final label -- whether the patient will experience tachyarrhythmia or not. In order to help the model learn better feature detectors in the hidden layers, we can simultaneously optimize it for additional related tasks. While the input layer and the first hidden layer are shared between all objectives, the upper part of the network branches off into three different prediction tasks. In addition to the main label, we train the network to also predict:

\begin{enumerate}
\itemsep0em
\item The New York Health Association Class (NYHAC), which indicates the severity of a patient’s functional impairment from congestive heart failure, ranging from class I-IV \cite{heartfailureclasses}.
\item The Body Mass Index (BMI) of the patient, quantifying the amount of tissue mass and indicating the level of obesity.
\end{enumerate}

While these patent-level values are available in the PhysioNet {\small MVTDB} dataset, they might not be available in a practical setting -- measuring the patient's BMI or assigning an NYHAC diagnosis is not always a priority in a critical care setting. By including them as additional objectives, the system uses the labels only during training and at test-time these values are not required. The performance improvement comes from jointly training the feature detectors in the first layer, along with the age embeddings, regularizing the network towards more robust internal representations.

\section{Experiments}

Experiments were performed on the PhysioNet {\small MVTDB} dataset \cite{goldberger2000physiobank}, collected from the records of 78 patients with ICDs (63 male and 15 female, aged 20.7 -- 75.3). The data contains 106 pre-VT records, 29 pre-VF records, and 126 control records, with 1024 RR intervals in each sequence prior to the corresponding event.
Previous work on VTA prediction has reported results using one third of the examples for testing and the rest for training. Since the dataset is relatively small and represents a low-resource scenario, this leaves very few examples for properly training the neural models and reporting reliable test results. Instead, we perform 10-fold cross-validation with stratified sampling over the dataset, which allows all of the datapoints to be used for training and evaluation. We also run each experiment 10 times with different random seeds and report the averaged results in this paper, in order to reduce the effect of possible outliers due to randomized initialization of the network weights.

Table \ref{tab:results} shows results of the different system configurations on the {\small MVTDB} dataset, predicting ventricular tachyarrhythmias 60 seconds in advance. The baseline model is a general multilayer network, using the 11 features chosen in previous work \cite{joo2012prediction}. This model achieves 62\% accuracy, which is lower than the previously reported 75\% -- this is expected, as the prediction time is now extended to 60 seconds instead of 10 seconds ahead of the episode. The next configuration changes the input features to the ones described in Sections \ref{sec:model} and \ref{sec:windowed}, introducing the window-based features and improving performance between 5-15\% across all metrics. Some of the features used in previous work have relatively low predictive power for this task and replacing them with metrics that can better capture the trend in the RR sequence proves to be very useful. Next, including the age embedding into the model provides another consistent improvement on all evaluation metrics. By using patient-level metadata, the model is able to make more presonalized decisions about the other features, indicating that the integration of more patient-specific features could be a promising direction for similar models.
Finally, training the model in a multi-task setting provides another further improvement on most metrics. By optimizing the network to predict different information about the patient, the system is motivated to learn more robust internal representations, which can be otherwise difficult to achieve with small datasets. The labels for NYHA Class and BMI value for the patient are not required during testing, therefore allowing the network to receive the benefit without any additional restrictions in practical applications.

\begin{table*}[t]
\centering
\setlength\tabcolsep{9pt}
\begin{tabular}{l|ccccc} \toprule
 & Accuracy & Sensitivity & Specificity & Precision & AUC \\ \midrule
Baseline & 62.26 & 63.19 & 61.27 & 63.60 & 68.55 \\
+ windowed features & 71.03 & 68.07 & 74.21 & 73.88 & 78.52 \\
+ age embedding & 73.52 & 70.07 & \textbf{77.22} & 76.74 & \textbf{79.65} \\
+ multi-task optimization & \textbf{74.02} & \textbf{71.04} & \textbf{77.22} & \textbf{76.99} & 78.77 \\ \bottomrule
\end{tabular}
\caption{System evaluation on the PhysioNet {\small MVTDB} dataset.}
\label{tab:results}
\end{table*}


\section{Conclusion}

We described a novel neural network architecture for the prediction of ventricular tachyarrhythmias. The model receives as input features that capture the direction and magnitude of the change in RR intervals and ectopic beats, along with several features based on heart rate variability and frequency analysis. 
Patient age is also included in the architecture as a trainable embedding, while the whole network is trained with multi-task optimization.
Each of these modifications provides a consistent improvement, achieving 74.02\% prediction accuracy and 77.22\% specificity 60 seconds in advance of the episode.
The model described in this paper could potentially be a useful component in a predictive monitoring system for improving the survival of patients at risk of imminent VTA. 


\bibliographystyle{unsrt}
\bibliography{references}

\end{document}